%% file: main.tex
\documentclass[10pt,twocolumn,letterpaper]{article}

\usepackage[pagenumbers]{cvpr}      

\input{preamble}

\definecolor{cvprblue}{rgb}{0.21,0.49,0.74}
\usepackage[pagebackref,breaklinks,colorlinks,citecolor=cvprblue]{hyperref}

\def\paperID{3122} 
\def\confName{CVPR}
\def\confYear{2024}

\title{Symphonize 3D Semantic Scene Completion with Contextual Instance Queries}

\def\name{Symphonies}

\author{
    Haoyi Jiang$^{1}$\thanks{Equal contribution. Work done during Haoyi Jiang's internship at Horizon Robotics.} \quad
    Tianheng Cheng$^{1*}$ \quad Naiyu Gao$^{2}$ \quad Haoyang Zhang$^{2}$ \quad Tianwei Lin$^{2}$ \\
    Wenyu Liu$^{1}$ \quad Xinggang Wang $^{1}$\thanks{Corresponding author.} \\
    $^1$ School of EIC, Huazhong University of Science \& Technology \\
    $^2$ Horizon Robotics \\
    {\tt\small \{haoyi\textunderscore jiang,thch,liuwy,xgwang\}@hust.edu.cn} \\
    {\tt\small \{naiyu.gao,haoyang.zhang,tianwei.lin\}@horizon.cc}
}

\begin{document}
\maketitle
\input{sec/0_abstract}
\input{sec/1_intro}
\input{sec/2_related_works}
\input{sec/3_method}
\input{sec/4_experiments}

\input{tab/8_sem_kitti_val}

\input{sec/5_conclusion}

\appendix
\section{Additional Experimental Results}
\label{sec:supp_expr}

\paragraph{Results on SemanticKITTI val.}

We present further quantitative results on the SemanticKITTI \texttt{val} set in \cref{tab:sem_kitti_val} for a more comprehensive comparison. \name{} consistently achieves state-of-the-art performance, aligning with our primary findings.

{
    \small
    \bibliographystyle{ieeenat_fullname}
    \bibliography{main}
}


\end{document}

%% file: preamble.tex
\usepackage{multirow}
\usepackage{graphicx}
\usepackage{wrapfig}
\usepackage[dvipsnames,table]{xcolor}

\usepackage{float}

\definecolor{White}{rgb}{1.,0.,1.}
\definecolor{first}{rgb}{.8,.0,.0}
\definecolor{second}{rgb}{.0,.6,.0}
\definecolor{third}{rgb}{.0,.0,.8}

\definecolor{ceiling}{RGB}{214,  38, 40}
\definecolor{floor}{RGB}{43, 160, 4}
\definecolor{wall}{RGB}{158, 216, 229}
\definecolor{window}{RGB}{114, 158, 206}
\definecolor{chair}{RGB}{204, 204, 91}
\definecolor{bed}{RGB}{255, 186, 119}
\definecolor{sofa}{RGB}{147, 102, 188}
\definecolor{table}{RGB}{30, 119, 181}
\definecolor{tvs}{RGB}{160, 188, 33}
\definecolor{furniture}{RGB}{255, 127, 12}
\definecolor{objects}{RGB}{196, 175, 214}

\definecolor{car}{rgb}{0.39215686, 0.58823529, 0.96078431}
\definecolor{bicycle}{rgb}{0.39215686, 0.90196078, 0.96078431}
\definecolor{motorcycle}{rgb}{0.11764706, 0.23529412, 0.58823529}
\definecolor{truck}{rgb}{0.31372549, 0.11764706, 0.70588235}
\definecolor{othervehicle}{rgb}{0.39215686, 0.31372549, 0.98039216}
\definecolor{person}{rgb}{1.        , 0.11764706, 0.11764706}
\definecolor{bicyclist}{rgb}{1.        , 0.15686275, 0.78431373}
\definecolor{motorcyclist}{rgb}{0.58823529, 0.11764706, 0.35294118}
\definecolor{road}{rgb}{1.        , 0.        , 1.        }
\definecolor{parking}{rgb}{1.        , 0.58823529, 1.        }
\definecolor{sidewalk}{rgb}{0.29411765, 0.        , 0.29411765}
\definecolor{otherground}{rgb}{0.68627451, 0.        , 0.29411765}
\definecolor{building}{rgb}{1.        , 0.78431373, 0.        }
\definecolor{fence}{rgb}{1.        , 0.47058824, 0.19607843}
\definecolor{vegetation}{rgb}{0.        , 0.68627451, 0.        }
\definecolor{trunk}{rgb}{0.52941176, 0.23529412, 0.        }
\definecolor{terrain}{rgb}{0.58823529, 0.94117647, 0.31372549}
\definecolor{pole}{rgb}{1.        , 0.94117647, 0.58823529}
\definecolor{trafficsign}{rgb}{1.        , 0.        , 0.        }
\definecolor{otherstructure}{rgb}{0.98039215, 0.58823529, 0.}
\definecolor{otherobject}{rgb}{0.19607843, 1.        , 1.        }

\makeatletter
\newcommand{\car@semkitfreq}{3.92}
\newcommand{\bicycle@semkitfreq}{0.03}
\newcommand{\motorcycle@semkitfreq}{0.03}
\newcommand{\truck@semkitfreq}{0.16}
\newcommand{\othervehicle@semkitfreq}{0.20}
\newcommand{\person@semkitfreq}{0.07}
\newcommand{\bicyclist@semkitfreq}{0.07}
\newcommand{\motorcyclist@semkitfreq}{0.05}
\newcommand{\road@semkitfreq}{15.30}
\newcommand{\parking@semkitfreq}{1.12}
\newcommand{\sidewalk@semkitfreq}{11.13}
\newcommand{\otherground@semkitfreq}{0.56}
\newcommand{\building@semkitfreq}{14.1}
\newcommand{\fence@semkitfreq}{3.90}
\newcommand{\vegetation@semkitfreq}{39.3}
\newcommand{\trunk@semkitfreq}{0.51}
\newcommand{\terrain@semkitfreq}{9.17}
\newcommand{\pole@semkitfreq}{0.29}
\newcommand{\trafficsign@semkitfreq}{0.08}
\newcommand{\semkitfreq}[1]{{\csname #1@semkitfreq\endcsname}}

\newcommand{\car@sscbkitfreq}{2.85}
\newcommand{\bicycle@sscbkitfreq}{0.01}
\newcommand{\motorcycle@sscbkitfreq}{0.01}
\newcommand{\truck@sscbkitfreq}{0.16}
\newcommand{\othervehicle@sscbkitfreq}{5.75}
\newcommand{\person@sscbkitfreq}{0.02}
\newcommand{\road@sscbkitfreq}{14.98}
\newcommand{\parking@sscbkitfreq}{2.31}
\newcommand{\sidewalk@sscbkitfreq}{6.43}
\newcommand{\otherground@sscbkitfreq}{2.05}
\newcommand{\building@sscbkitfreq}{15.67}
\newcommand{\fence@sscbkitfreq}{0.96}
\newcommand{\vegetation@sscbkitfreq}{41.99}
\newcommand{\terrain@sscbkitfreq}{7.10}
\newcommand{\pole@sscbkitfreq}{0.22}
\newcommand{\trafficsign@sscbkitfreq}{0.06}
\newcommand{\otherstructure@sscbkitfreq}{4.33}
\newcommand{\otherobject@sscbkitfreq}{0.28}
\newcommand{\sscbkitfreq}[1]{{\csname #1@sscbkitfreq\endcsname}}

%% file: sec/0_abstract.tex
\begin{abstract}
    3D Semantic Scene Completion (SSC) has emerged as a nascent and pivotal undertaking in autonomous driving, aiming to predict voxel occupancy within volumetric scenes.
    However, prevailing methodologies primarily focus on voxel-wise feature aggregation, while neglecting instance semantics and scene context.
    In this paper, we present a novel paradigm termed \textbf{\textit{Symphonies (Scene-from-Insts)}}, that delves into the integration of instance queries to orchestrate 2D-to-3D reconstruction and 3D scene modeling.
    Leveraging our proposed Serial Instance-Propagated Attentions, Symphonies dynamically encodes instance-centric semantics, facilitating intricate interactions between image-based and volumetric domains.
    Simultaneously, Symphonies enables holistic scene comprehension by capturing context through the efficient fusion of instance queries, alleviating geometric ambiguity such as occlusion and perspective errors through contextual scene reasoning.
    Experimental results demonstrate that Symphonies achieves state-of-the-art performance on challenging benchmarks—SemanticKITTI and SSCBench-KITTI-360, yielding remarkable mIoU scores of 15.04 and 18.58, respectively. These results showcase the paradigm's promising advancements.
    The code for our method is available at \url{https://github.com/hustvl/Symphonies}.
\end{abstract}

%% file: sec/1_intro.tex
\section{Introduction}
\label{sec:intro}

The advent of autonomous driving has brought forth novel challenges in the realm of 3D perception. In the pursuit of safe navigation and obstacle avoidance, autonomous vehicles must possess the capability to accurately predict the occupancy of their immediate surroundings.
This task, however, is not a facile endeavor, given the inherent complexities of the real world, characterized by clutter, ambiguity, and rapid evolution.

3D Semantic Scene Completion (SSC) formulates this challenge as the reconstruction of both occupancy and semantics for every volumetric grid within a 3D scene. Recent advancements in vision-based solutions, such as MonoScene~\cite{MonoScene} and OccDepth~\cite{OccDepth}, adopt 3D convolutional networks to elevate 2D image features into the 3D domain. TPVFormer~\cite{TPVFormer}, OccFormer~\cite{OccFormer}, and CTF-Occ~\cite{Occ3D} explore decomposing 3D volumes into coarse view representations and enhancing voxel interactions using Transformer~\cite{Transformer, DeformableDETR, BEVFormer} architectures.

Despite these advancements, contemporary approaches tend to prioritize voxel-wise modeling for 3D scenes and resort to pixel-voxel projection~\cite{OFT, LSS, BEVFormer, GKT} for dimension promotion. While focusing on these regional representations, they inadvertently neglect higher-level instance semantics, leading to vulnerability to geometric ambiguity arising from occlusion and perspective errors.
Humans, in contrast, naturally perceive and comprehend through the concept of ``instance'', where segments of pixels or voxels provide valuable semantic insights, while collectively shaping the scene's context. In light of these limitations, a fundamental question arises: \textbf{\textit{How can we leverage the notion of instances to steer 3D scene modeling and 2D-to-3D reconstruction?}}

\begin{figure*}
    \centering
    \includegraphics[width=0.7\linewidth ]{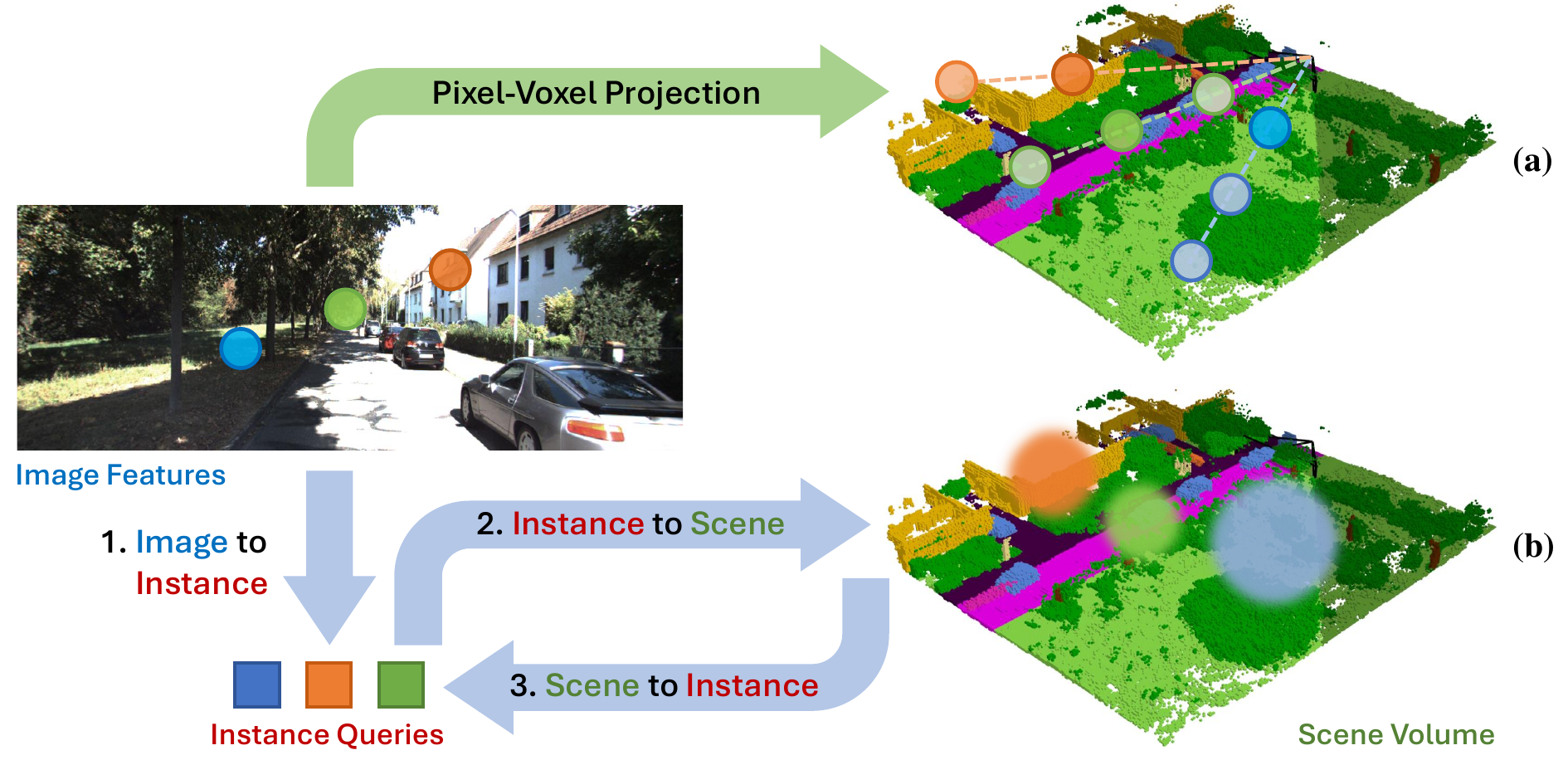}
    \caption{\textbf{Comparison between voxel-wise modeling (a) and Symphonies (b).} Conventional methods primarily depend on Inverse Perspective Mapping (IPM)-based voxel-pixel projection and voxel-wise feature aggregation, resulting in geometric ambiguity and computational redundancy. In contrast, Symphonies leverages instance queries as intermediaries to engage with image and voxel features, thus exploiting instance semantics and enhancing the contextual comprehension of the scene.}
    \label{fig:teaser}
\end{figure*}

Drawing inspiration from this notion, we propose Symphonies (Scene-from-Insts), a novel method that leverages contextual instance queries derived from image inputs to enhance scene modeling, exploiting inherent instance semantics and scene context.
Stemming from this basis, we propose Serial Instance-Propagated Attentions to intricately interact with image and voxel features, deformably aggregating instance-centric semantics. This seamless interaction bridges the gap between low-level pixel/voxel representations and high-level semantics, facilitating dimension promotion and scene modeling, as illustrated in \cref{fig:teaser}.
Furthermore, the fusion of multiple instance queries collectively enriches broader contextual information for scene reasoning, contributing to the alleviation of geometric ambiguity.
In tandem, we introduce the Depth-Rectified Voxel Proposal Layer to refine the initial geometry, elevating 2D image features to the implicit surface of the scene.

To evaluate the effectiveness of our method, extensive experiments are conducted on the challenging SemanticKITTI~\cite{SemanticKITTI} and SSCBench-KITTI-360~\cite{KITTI360, SSCBench} datasets. Symphonies achieves a remarkable state-of-the-art performance of 15.04 and 18.58 mIoU, respectively, significantly outperforming previous vision-based methods by a substantial margin. Ablation experiments further underscore the promising advancements of our approach in the field of SSC.
In summary, our contributions involve:

\begin{itemize}
     \item We introduce Symphonies, a pioneering paradigm for 3D Semantic Scene Completion (SSC), which delves into modeling instance-centric semantics using sparse instance queries, facilitating efficient interactions between image and volume features through our proposed Serial Instance-Propagated Attentions.
     \item Symphonies effectively captures global scene context through the fusion of instance queries, enabling a holistic comprehension of the surroundings. The scene context contributes to mitigating geometric ambiguity via contextual scene reasoning and refined geometry provided by the proposed Depth-Rectified Voxel Proposal Layer.
     \item Our proposed method significantly surpasses existing approaches on challenging SSC benchmarks, achieving 15.04 mIoU on SemanticKITTI and 18.58 mIoU on SSCBench-KITTI-360. These results underscore the considerable potential of our paradigm in advancing autonomous driving and scene understanding.
\end{itemize}

%% file: sec/2_related_works.tex
\section{Related Works}
\label{sec:related_works}

\begin{figure*}
    \centering
    \includegraphics[width=\linewidth]{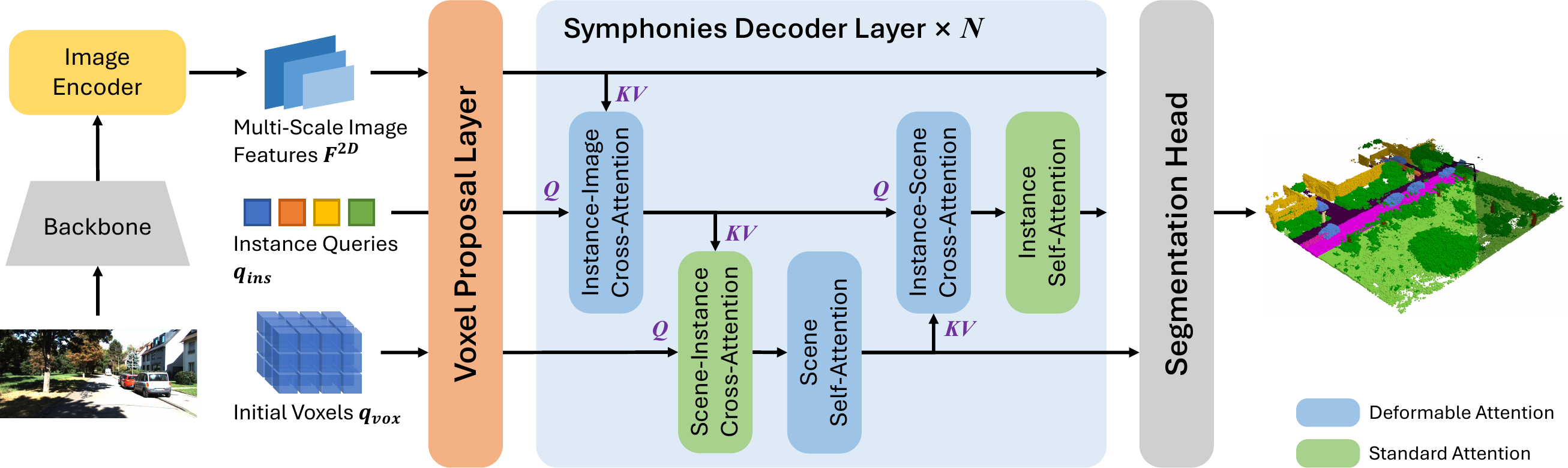}
    \caption{\textbf{Overview of Symphonies.} The Symphonies framework encompasses several key components. It commences with extracting multi-scale image features via the image backbone and Instance-Aware Image Encoder. The Depth-Rectified Voxel Proposal Layer generates initial voxel features estimating the implicit surface. Subsequently, the Symphonies Decoder Layers, which consist of Serial Instance-Propagated Attentions, facilitate continuous interactions among the image, instances, and the scene, iterated $N$ times. The Segmentation Head upsamples voxel features to the designated resolution and predicts class logits for each voxel.}
    \label{fig:main_arch}
\end{figure*}

\paragraph{3D Semantic Scene Completion.}
3D Semantic Scene Completion (SSC) entails predicting occupancy and semantics for each voxel within a 3D scene, which was initially introduced by SSCNet~\cite{SSCNet}. Subsequent methods can be broadly categorized based on their model architectures and input modalities.
Volume networks~\cite{DDRNet, CCPNet} predominantly utilize Truncated Signed Distance Function (TSDF) features generated from depth data, processed through 3D convolutional networks. On the other hand, view-volume networks~\cite{VVNet, SATNet, ForkNet, LMSCNet, AICNet} extract RGB or depth features with 2D networks before converting them into 3D volumes. For a more in-depth overview of SSC, we refer readers to the survey by Rold{\~{a}}o \etal~\cite{SSCSurvey}.

Recently, camera-based SSC has garnered increasing attention for its immense potential in the field of autonomous driving. MonoScene~\cite{MonoScene} presents the first purely visual solution, sampling RGB features along the line of sight and adapting a 3D UNet architecture. TPVFormer~\cite{TPVFormer} introduces a Tri-Perspective View (TPV) representation to decompose voxels onto various view planes for efficient scene encoding. VoxFormer~\cite{VoxFormer} proposes a two-stage framework that diffuses the global scene from proposed voxel features, resembling the Masked Autoencoder (MAE)~\cite{MAE}. OccDepth~\cite{OccDepth} improves 2D-to-3D geometric projection leveraging implicit stereo depth information. OccFormer~\cite{OccFormer} applies a Transformer-based decoder and a mask-wise prediction paradigm akin to MaskFormer~\cite{MaskFormer, Mask2Former}. NDC-Scene~\cite{NDCScene} explores alleviating geometric ambiguity through Normalized Device Coordinates (NDC). OccNet~\cite{OccNet} further envisions occupancy as a general scene descriptor for a wide scope of driving tasks.

In contrast to prior works, our proposed Symphonies differs by integrating instance queries to enhance scene modeling through instance semantics and enriched contextual awareness, mitigating geometric ambiguity arising from voxel-wise modeling without resorting to additional Bird's Eye View (BEV) or occupancy prediction networks.

\paragraph{Camera-Based 3D Perception.}
The surge in autonomous driving applications has rekindled interest in camera-based 3D perception, given its cost-effectiveness and alignment with human visual perception.
Early 3D object detection methods, such as FCOS3D~\cite{FCOS3D} and DETR3D~\cite{DETR3D}, straightforwardly extend 2D detectors to predict additional 3D bounding boxes. Among subsequent Transformer-based approaches, BEVFormer~\cite{BEVFormer} and BEVDet~\cite{BEVDet} adopt the BEV space to align multi-frame features, while PolarDETR~\cite{PolarDETR} establishes explicit correlations between image patterns. In addition, PETR~\cite{PETR} and PETRv2~\cite{PETRv2} utilize 3D position embeddings to encode 2D features.

BEV segmentation, which is beneficial for representation learning and route planning, has also been explored. Approaches such as OFT~\cite{OFT}, Lift-Splat~\cite{LSS}, and FIERY~\cite{FIERY} transform the camera plane into BEV via Inverse Perspective Mapping (IPM).
PolarBEV~\cite{PolarBEV} uses angle-specific and radius-specific embeddings to rasterize BEV features. BEVFormer~\cite{BEVFormer} and CVT~\cite{CVT} aggregate BEV queries through cross-attention layers, while GKT~\cite{GKT} optimizes computational efficiency by constraining local attention calculations.

These aspects closely relate to our work in SSC, where techniques like Deformable Attention~\cite{DeformableDETR} inspire our methodology to enhance 3D scene completion.

%% file: sec/3_method.tex
\section{Scene from Instances}

This section presents a comprehensive elaboration of our proposed Symphonies method, beginning with an architectural overview in \cref{sec:overview}. It proceeds to detail the Depth-Rectified Voxel Proposal Layer in \cref{sec:voxel_proposal} and the Symphonies Decoder Layer in \cref{sec:symphonies_layer}, shedding light on their synergistic contributions. Further insights into training losses are discussed in \cref{sec:losses}.

\subsection{Overview}
\label{sec:overview}

The architectural details of our proposed Symphonies are illustrated in \cref{fig:main_arch}. In essence, Symphonies exclusively takes RGB images as input and extracts multi-scale 2D features $F^{2D}$ through a ResNet-50~\cite{ResNet} image backbone and an Instance-Aware Deformable Transformer~\cite{DeformableDETR} Encoder, enhancing both global and instance semantics on the image plane.
In the Symphonies Decoder, instance queries $q_{ins} \in \mathbb{R}^{N \times C}$ and the volumetric scene representation $q_{vox} \in \mathbb{R}^{C \times X \times Y \times Z}$ are initialized with learnable embeddings. Here, $C$ signifies embedding dimensions, $N$ denotes the number of instance queries, while $X$, $Y$, and $Z$ indicate the scene grid dimensions.

The subsequent ``scene-from-instances'' process commences with the Depth-Rectified Voxel Proposal Layer initializing voxel proposals $q_p$ with image features on the implicit surface. Multi-scale image features $F^{2D}$, scene features $q_{vox}$, and instance queries $q_{ins}$ are passed through our proposed Serial Instance-Propagated Attentions within the Symphonies Decoder Layers. This iterative process continuously propagates image features $F^{2D}$ to scene features $q_{vox}$ guided by instance queries $q_{ins}$, while simultaneously aggregating instance semantics from both modalities.
The Segmentation Head then upsamples the scene features to the target resolution, and predicts per-voxel class logits with a single $1 \times 1 \times 1$ convolution after an Atrous Spatial Pyramid Pooling (ASPP)~\cite{ASPP} module.

\paragraph{Depth Estimator.}
The depth prediction, acquired from a pre-trained depth estimator, is not explicitly illustrated in the diagram for clarity. It is employed to infer the implicit surface within the Voxel Proposal Layer and compute instance reference points in the scene volume. Specifically, we adopt the pre-trained Mobilestereonet~\cite{MobileStereoNet} as the depth estimator, aligning with VoxFormer~\cite{VoxFormer}.

\paragraph{Instance-Aware Image Encoder.}
The Instance-Aware Image Encoder, vital for integrating instance semantics in the absence of direct instance-level supervision, employs a Deformable Transformer~\cite{DeformableDETR} adept at capturing long-range dependencies around diverse instances by attending to deformable reference points.
Additionally, it is augmented by utilizing the pre-trained weight of MaskDINO~\cite{MaskDINO} from panoptic segmentation~\cite{PanopticFPN}, to enrich its instance awareness.

\subsection{Depth-Rectified Voxel Proposal Layer}
\label{sec:voxel_proposal}

The Depth-Rectified Voxel Proposal Layer generates initial scene features for voxels located on the implicit surface, known as voxel proposals, which establishes coarse geometry awareness for subsequent instance-level aggregations. The implicit surface is computed through the conversion of camera coordinates to world coordinates using depth estimation, described as follows:
\begin{align}
    x^C &= \ \mathcal{K}^{-1} \cdot (z_c \odot x^I) \\
    x^W &= \ [R, T]^{-1} \cdot x^C
\end{align}
where $x^I$, $x^C$, and $x^W$ represent homogeneous coordinates of pixels, camera frustum, and the world, respectively. $\odot$ denotes the element-wise multiplication. The intrinsic matrix $\mathcal{K}$ encompasses camera parameters, while the extrinsic matrix is composed of the rotation matrix $R$ and the translation vector $T$. $z_c$ corresponds to the z-coordinate of the camera, \ie, the depth estimation.

Based on the camera-to-world transformation, the positions $V_p$ of voxel proposals are determined by mapping image points $x^I$ to their corresponding world coordinates $x^W$, confined within the volume $V$:
\begin{align}
    V_p = \{x^W \mid & \ x^W = \mathcal{T}^{IW}(x^I, z_c), \notag \\
                     & \ \forall \ x^I \in I \ \text{such that} \ x^W \in V\}
\end{align}
Here, $\mathcal{T}^{IW}$ refers to the camera-to-world transformation, $I$ represents image pixels, and $V$ represents voxel grids.

\begin{figure}
    \centering
    \includegraphics[width=0.4\textwidth]{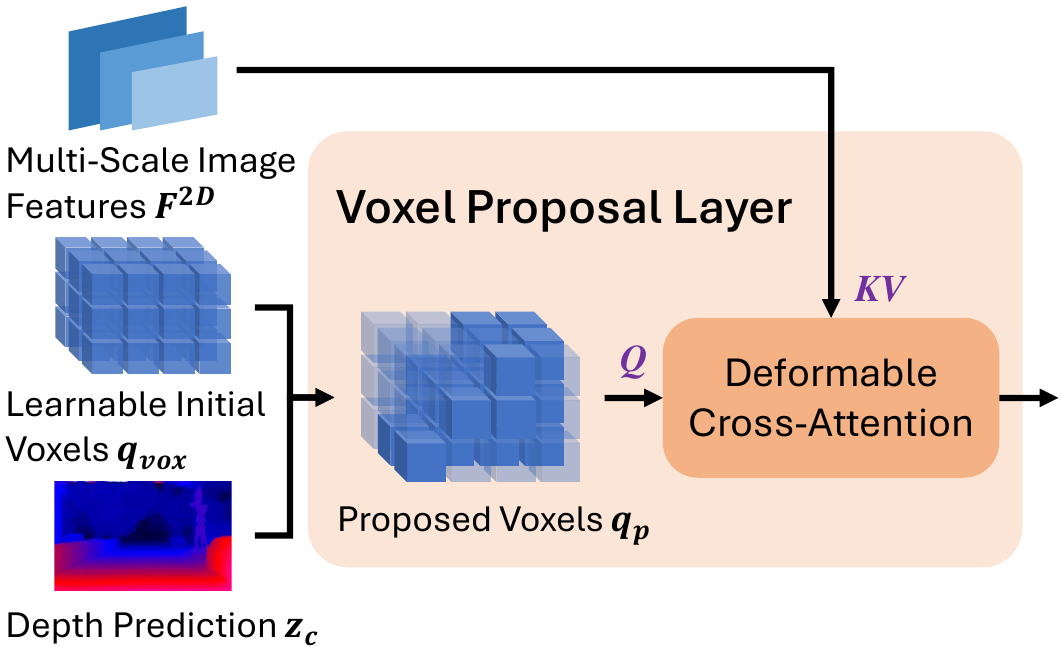}
    \caption{\textbf{Illustration of the Depth-Rectified Voxel Proposal Layer.}}
    \label{fig:vpl}
\end{figure}

As illustrated in \cref{fig:vpl}, the determined voxel features are initialized by aggregating multi-scale image features using Deformable Attention~\cite{DeformableDETR}. This process involves selecting the proposed voxels $q_p$ associated with the positions $V_p$ from scene volume $q_{vox}$, along with corresponding pixel positions $p_I$ and 2D image features $F^{2D}$. This process is expressed as $q_p = \text{DeformAttn}(q_p, p_I, F^{2D})$.

The Deformable Attention operation, denoted as $\texttt{DeformAttn}$, dynamically aggregates query features $q$ from features $x$ with deformable reference points $p_q$. The mathematical expression is given by:
\begin{equation}
    \text{DeformAttn}(q, p_q, x) = \sum_{k=1}^{K} A_{qk} W x(p_q + \Delta p_{qk})
\end{equation}
Here, $K$ represents the number of sampling points, and $A_{qk}$ stands for the learnable attention weight at sampling point $k$ deformable based on queries $q$. The term $\Delta p_{qk}$ denotes the offset applied to $p_q$, and $W$ denotes the projection weight. The computation of multi-heads is omitted for brevity.

In contrast to the Query Proposal in VoxFormer~\cite{VoxFormer}, which employs an extra occupancy network~\cite{LMSCNet} for generating coarse occupancy features, we refrain from it as it introduces additional geometric ambiguity in occlusion regions.

\subsection{Symphonies Decoder Layer}
\label{sec:symphonies_layer}

The Symphonies Decoder Layer orchestrates continuous interactions between instance queries and scene representations, exploiting the instance-centric semantics overlooked in prior voxel-wise modeling methods, thereby enhancing 3D scene modeling. It consists of the proposed Serial Instance-Propagated Attentions, as shown in \cref{fig:main_arch}.
Firstly, the deformable cross-attention modules for instance-image and instance-scene interactions enable instance queries to selectively attend to relevant segments from both modalities, aggregating instance semantics.
Furthermore, the instance self-attention module strengthens the internal structures of semantically enriched instance queries while collectively synthesizing contextual information for holistic scene comprehension.
Subsequently, the scene-instance cross-attention module aggregates scene features from instance queries, leveraging instance semantics and global context, mitigating geometric ambiguity through context-aware reasoning.
The scene self-attention mechanism further diffuses voxel features throughout the entire scene.

The following paragraphs present a detailed explanation of the computations involved in their exact order of operation. To streamline the explanation, detailed discussions on certain components, including Feed-Forward Networks (FFN), Layer Norms (LN), and identity connections, have been omitted.

\paragraph{Deformable Instance-Image Cross-Attention.}
For each instance query $q_{ins}$, deformable attention extracts surrounding features from multi-scale image features $F^{2D}$ using learnable 2D reference points $p_{ins}^{2D}$, denoted as $q_{ins} = \text{DeformAttn}(q_{ins}, p_{ins}^{2D}, F^{2D})$.

\paragraph{Scene-Instance Cross-Attention.}
This attention mechanism aggregates scene features $q_{vox}$ from instance queries, formulated as $q_{vox}^{\in FOV} = \text{CrossAttn}(q_{vox}^{\in FOV}, q_{ins}, q_{ins})$, where $FOV$ refers to the ``field of view'' which is pre-computed based on world-to-camera transformation excluding invisible voxels, reducing computational redundancy.

\paragraph{Deformable Scene Self-Attention.}
The scene self-attention enables feature propagation across the scene, where voxels attend to their neighbors: $q_{vox}^{\in FOV} = \text{DeformAttn}(q_{vox}^{\in FOV}, p_{V}, q_{vox})$. Here, $p_{V}$ represents voxels' relative coordinates in the scene.

\paragraph{Deformable Instance-Scene Cross-Attention.}
Instance semantics are enhanced by integrating refined information from the reconstructed voxel features $q_{vox}$. Through the transformation applied to 2D reference points, 3D reference points are derived as $p_{ins}^{3D} = \mathcal{T}^{IW}(p_{ins}^{2D})$. The instance-scene cross-attention are then formulated as $q_{ins} = \text{DeformAttn}(q_{ins}, p_{ins}^{3D}, q_{vox})$.

\paragraph{Instance Self-Attention.}
The instance self-attention captures internal relations and global context within instance queries, expressed as $q_{ins} = \text{SelfAttn}(q_{ins})$.

\subsection{Losses}
\label{sec:losses}

In the Symphonies framework, we adopt the Scene-Class Affinity Loss $L_{scal}$ from MonoScene~\cite{MonoScene} to optimize precision, recall, and specificity concurrently. A detailed description of this loss is provided in the supplementary material. The Scene-Class Affinity Loss is applied to semantic and geometric predictions, in conjunction with the cross-entropy loss weighted by class frequencies. The overall loss function is formulated as follows:
\begin{equation}
    \mathcal{L} = \mathcal{L}_{scal}^{geo} + \mathcal{L}_{scal}^{sem} + \mathcal{L}_{ce}
\end{equation}

Following the DETR series~\cite{DETR}, auxiliary losses are applied after each Symphonies Decoder Layer for enhanced supervision, following the same formulation as $\mathcal{L}$ but scaled by a factor of 0.5.

%% file: sec/4_experiments.tex
\section{Experiments}
\label{sec:experiments}

\input{tab/1_sem_kitti}
\input{tab/2_kitti_360}

In this section, we present the evaluation results of our proposed \name{} on SemanticKITTI~\cite{SemanticKITTI} and SSCBench-KITTI-360~\cite{SSCBench} datasets. The comparative analysis including the performance of \name{} against existing approaches is detailed in \cref{sec:main_results}. Additionally, comprehensive ablation studies are conducted in \cref{sec:ablation} to shed light on the thorough understanding of \name.

\subsection{Dataset and Metric}
The evaluation is performed on SemanticKITTI~\cite{SemanticKITTI} and SSCBench-KITTI-360~\cite{SSCBench} datasets, both providing densely annotated urban driving scene sequences, 22 and 9 respectively, from the KITTI Odometry Benchmark~\cite{KITTI}. These datasets voxelize the point clouds and label the entire scene measuring $51.2m\times51.2m\times64m$, with voxel grids of $256\times256\times32$ and voxel size of 0.2m.
SemanticKITTI comprises 10 sequences for training, 1 sequence for validation, and 11 sequences for testing. It furnishes RGB images with shapes of $1226 \times 370$ as inputs and encompasses 20 semantic classes. SSCBench-KITTI-360 provides 7 sequences for training, 1 sequence for validation, and 1 sequence for testing, with 19 semantic classes, where RGB images have a resolution of $1408 \times 376$ as inputs.
For our camera-based approach, we exclusively adopt RGB images as input, and report the intersection over union (IoU) and mean IoU (mIoU) metrics for occupied voxel grids and voxel-wise semantic predictions respectively, aligned with standard practices.

\subsection{Implementation Details}
In line with prior studies~\cite{MonoScene, TPVFormer, VoxFormer}, we train \name{} for 30 epochs on 4 NVIDIA 3090 GPUs, with a batch size of 4 images. We apply random horizontal flip augmentation and employ the AdamW~\cite{AdamW} optimizer with an initial learning rate of 2e-4 and a weight decay of 1e-4. Learning rate reduction occurs by a factor of 0.1 at the 25th epoch.
The ResNet-50~\cite{ResNet} backbone and Image Encoder are initialized with pre-trained MaskDINO~\cite{MaskDINO} weights.

\subsection{Main Results}
\label{sec:main_results}
We conduct a comprehensive comparison of \name{} with the latest state-of-the-art camera-based methodologies on the SemanticKITTI and SSCBench-KITTI-360 datasets. The results, outlined in \cref{tab:sem_kitti_test} and \cref{tab:kitti_360_test}, establish the superior performance of \name{}. It exhibits substantial improvements of 2.72 and 4.77 mIoU on SemanticKITTI and SSCBench-KITTI-360, respectively.
\name{} showcases particular excellence in instance classes, \eg, buildings, cars, persons, and bicycles. This underscores its prowess in capturing and modeling intricate instance semantics.
While VoxFormer attains a marginally higher IoU on SemanticKITTI, its adoption of two-stage training and extra coarse occupancy prediction network disrupts end-to-end training and introduces additional geometric ambiguity. This complexity hampers its robustness, especially on KITTI-360.

The superiority of \name{} becomes more pronounced on the SSCBench-KITTI-360 benchmark, where it surpasses other camera-based counterparts by a substantial margin of 4.77 mIoU, attributed to the ample data samples and high-quality annotations. Moreover, \name{} even outperforms LiDAR-based methods in terms of mIoU, despite LiDAR's inherent advantage in IoU owing to its more precise position awareness, particularly at a distance.

\input{tab/3_ablat_mods}

\subsection{Ablation Studies}
\label{sec:ablation}

The ablation analysis is conducted on the SemanticKITTI validation set from four key perspectives: overall architectural components, the Symphonies Decoder, the Voxel Proposal Layer, and the Image Encoder.

\paragraph{Ablation on architectural components.}
\cref{tab:ablat_mods} presents the breakdown analysis of various architectural components within \name{}. Commencing with a ResNet-50 backbone, an Image Encoder without pre-trained weight, a 2D-to-3D projection via FLoSP~\cite{MonoScene}, and a single 3D ASPP layer as the 3D decoder, the vanilla baseline can be considered as a light-weight alternative to MonoScene.
Initializing the Image Encoder with pre-trained weights leads to a notable improvement of 2.15 mIoU, emphasizing the effectiveness of instance awareness brought by 2D segmentation pre-training.
Further, the proposed Depth-Rectified Voxel Proposal Layer improves performance by 0.75 mIoU through more accurate geometry. 
The Symphonies Decoder significantly boosts performance by 5.38 IoU, attributed to its dynamic instance modeling and context-capturing capabilities.
In summary, the analysis in \cref{tab:ablat_mods} affirms the effectiveness of the proposed components in \name{}.

\paragraph{Ablation on the Symphonies Decoder.}
To gain insights into the functionality of contextual instance queries, we assess the modular interactions within the \name{} Decoder Layer.
As depicted in \cref{tab:ablat_attns}, the incorporation of instance queries with either instance-image or instance-scene cross-attention considerably enhances performance. This substantiates the significance of instance queries for adaptive aggregation of instance semantics.
Among them, the instance-image cross-attention brings less improvement, suggesting that original image features have already been adequately captured in the preceding Voxel Proposal Layer. Furthermore, instance self-attention further yields an improvement of 0.52 IoU, underlining the contextual effectiveness of efficient fusion among instance queries.

\input{tab/4_ablat_decoder}

\paragraph{Ablation on the Voxel Proposal Layer.}
Comparing the Depth-Rectified Voxel Proposal Layer (VPL) with FLoSP from MonoScene~\cite{MonoScene} casting pixels to voxels along the line of sight, as well as the mono VPL using monocular depth estimator AdaBins~\cite{AdaBins} (0.058 REL on KITTI), we note significant occupancy prediction improvements using the stereo VPL based on MobileStereoNet~\cite{MobileStereoNet} (0.66 EPE on KITTI 2015), as shown in \cref{tab:ablat_vpl}. This indicates that the rectification of more precise depth estimation contributes to mitigating geometric ambiguity, aligning with the findings in VoxFormer.

\input{tab/5_ablat_vpl}

\paragraph{Ablation on the Instance-Aware Image Encoder.}
\cref{tab:ablat_encoder} evidently showcases the synergistic effects of utilizing pre-trained weights for the Image Encoder with instance queries.
Solely utilizing pre-trained weights from MaskDINO~\cite{MaskDINO} contributes an additional improvement of 0.48 mIoU. Moreover, incorporating instance queries with the pre-trained encoder yields a significant improvement of 1.36 mIoU, implying that the proposed instance queries benefit from the enhanced instance awareness of the encoder.

\input{tab/6_ablat_encoder}

\begin{figure*}
    \centering
    \includegraphics[width=0.8\linewidth ]{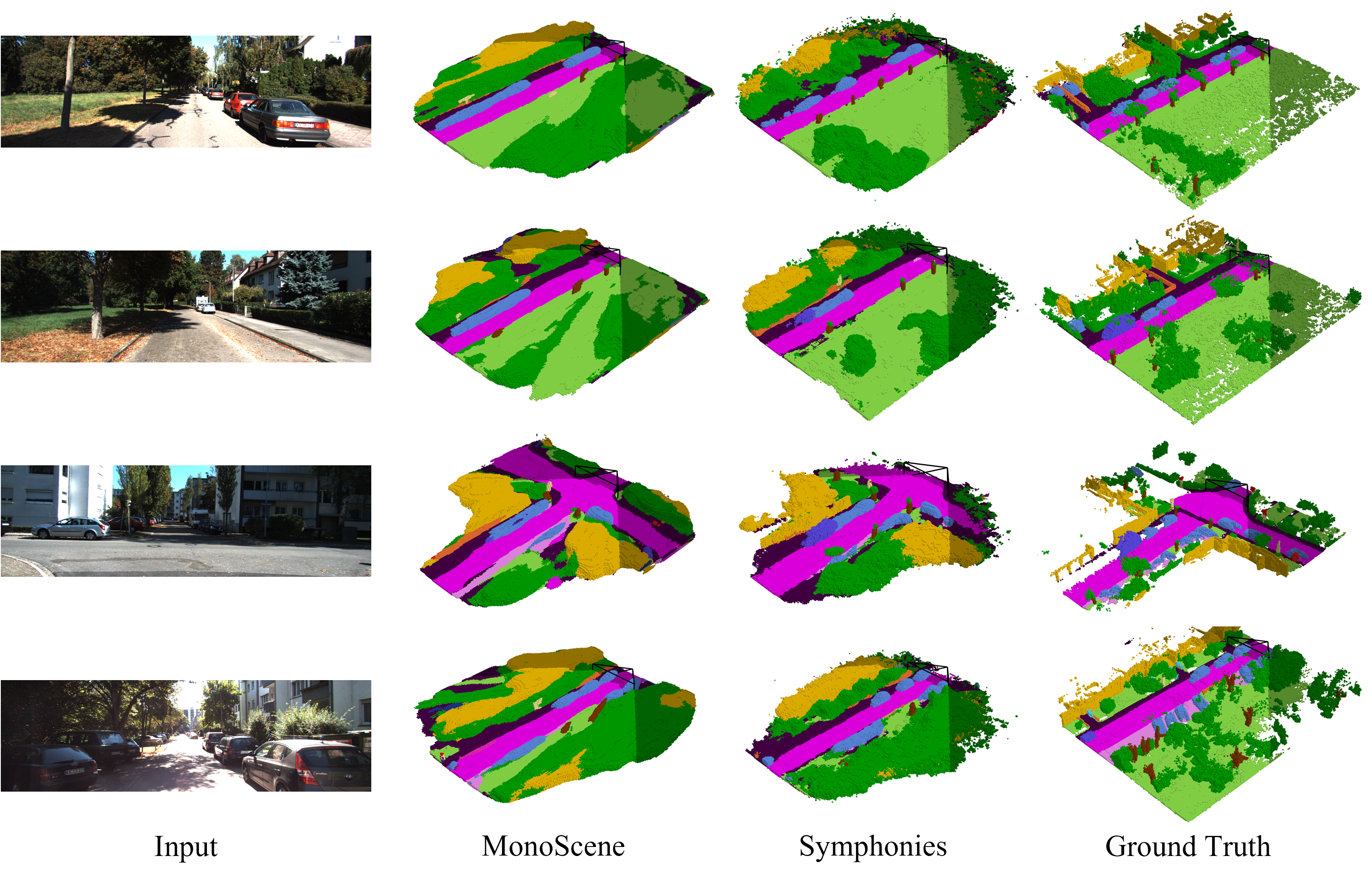}
    \caption{\textbf{Qualitative visualizations on SemanticKITTI \texttt{val}.} Symphonies consistently produces detailed predictions for objects such as cars and trunks, while maintaining coherent layouts for structures like buildings and vegetation.}
    \label{fig:vis}
\end{figure*}

\subsection{Visualizations}

\paragraph{Qualitative Results.} \cref{fig:vis} presents the visualizations of Symphonies on SemanticKITTI \texttt{val}, in comparison to the counterpart MonoScene.
Symphonies generates more detailed predictions for instance-centric classes such as cars and trunks, as well as preserves clear and coherent layouts for structures like buildings and vegetation, attributed to the enriched instance semantics and contextual information provided by instance queries.
In contrast, MonoScene produces vague predictions with a radial shape, which is indicative of the aforementioned ambiguous geometry. These results underscore the superior capability of Symphonies in capturing fine-grained scene representations and enhancing overall scene understanding.

\paragraph{Attention Map Analysis.} The attention map analysis in \cref{fig:vis_attn} provides insights into the mechanisms of the Serial Instance-Propagated Attentions within \name{} layers. Notably, instance queries exhibit selective attention to corresponding regions in both the image and the scene. Additionally, they activate the semantically related regions within the scene-instance cross-attention. This observation validates the effect of our claimed instance-centric semantics in facilitating effective scene modeling.

%% file: tab/1_sem_kitti.tex
\begin{table*}[ht]
    \centering
    \newcommand{\clsname}[2]{
        \rotatebox{90}{
            \hspace{-6pt}
            \textcolor{#2}{$\blacksquare$}
            \hspace{-6pt}
            \renewcommand\arraystretch{0.6}
            \begin{tabular}{l}
                #1                                      \\
                \hspace{-4pt} ~\tiny(\semkitfreq{#2}\%) \\
            \end{tabular}
        }}
    \renewcommand{\tabcolsep}{2pt}
    \renewcommand\arraystretch{1.1}
    \scalebox{0.84}
    {
        \begin{tabular}{l|r>{\columncolor{gray!20}}r|rrrrrrrrrrrrrrrrrrrr}
            \toprule
            Method                               &
            \multicolumn{1}{c}{IoU}              &
            mIoU                                 &
            \clsname{road}{road}                 &
            \clsname{sidewalk}{sidewalk}         &
            \clsname{parking}{parking}           &
            \clsname{other-grnd.}{otherground}   &
            \clsname{building}{building}         &
            \clsname{car}{car}                   &
            \clsname{truck}{truck}               &
            \clsname{bicycle}{bicycle}           &
            \clsname{motorcycle}{motorcycle}     &
            \clsname{other-veh.}{othervehicle}   &
            \clsname{vegetation}{vegetation}     &
            \clsname{trunk}{trunk}               &
            \clsname{terrain}{terrain}           &
            \clsname{person}{person}             &
            \clsname{bicyclist}{bicyclist}       &
            \clsname{motorcyclist}{motorcyclist} &
            \clsname{fence}{fence}               &
            \clsname{pole}{pole}                 &
            \clsname{traf.-sign}{trafficsign}
            \\
            \midrule
            LMSCNet$^\dagger$~\cite{LMSCNet}     & 31.38          & 7.07           & 46.70          & 19.50          & 13.50          & 3.10           & 10.30          & 14.30          & 0.30          & 0.00          & 0.00          & 0.00          & 10.80          & 0.00           & 10.40          & 0.00          & 0.00          & 0.00          & 5.40           & 0.00          & 0.00          \\
            AICNet$^\dagger$~\cite{AICNet}       & 23.93          & 7.09           & 39.30          & 18.30          & 19.80          & 1.60           & 9.60           & 15.30          & 0.70          & 0.00          & 0.00          & 0.00          & 9.60           & 1.90           & 13.50          & 0.00          & 0.00          & 0.00          & 5.00           & 0.10          & 0.00          \\
            JS3C-Net$^\dagger$~\cite{JS3CNet}    & 34.00          & 8.97           & 47.30          & 21.70          & 19.90          & 2.80           & 12.70          & 20.10          & 0.80          & 0.00          & 0.00          & 4.10          & 14.20          & 3.10           & 12.40          & 0.00          & 0.20          & 0.20          & 8.70           & 1.90          & 0.30          \\
            MonoScene$^\ast$~\cite{MonoScene}    & 34.16          & 11.08          & 54.70          & 27.10          & 24.80          & 5.70           & 14.40          & 18.80          & 3.30          & 0.50          & 0.70          & 4.40          & 14.90          & 2.40           & 19.50          & 1.00          & 1.40          & 0.40          & 11.10          & 3.30          & 2.10          \\
            TPVFormer~\cite{TPVFormer}           & 34.25          & 11.26          & 55.10          & 27.20          & 27.40          & 6.50           & 14.80          & 19.20          & \textbf{3.70} & 1.00          & 0.50          & 2.30          & 13.90          & 2.60           & 20.40          & 1.10          & 2.40          & 0.30          & 11.00          & 2.90          & 1.50          \\
            VoxFormer~\cite{VoxFormer}           & \textbf{42.95} & 12.20          & 53.90          & 25.30          & 21.10          & 5.60           & 19.80          & 20.80          & 3.50          & 1.00          & 0.70          & 3.70          & 22.40          & 7.50           & 21.30          & 1.40          & \textbf{2.60} & 0.20          & 11.10          & 5.10          & 4.90          \\
            OccFormer~\cite{OccFormer}           & 34.53          & 12.32          & 55.90          & \textbf{30.30} & \textbf{31.50} & 6.50           & 15.70          & 21.60          & 1.20          & 1.50          & 1.70          & 3.20          & 16.80          & 3.90           & 21.30          & 2.20          & 1.10          & 0.20          & 11.90          & 3.80          & 3.70          \\
            \hline
            \name{}                              & 42.19          & \textbf{15.04} & \textbf{58.40} & 29.30          & 26.90          & \textbf{11.70} & \textbf{24.70} & \textbf{23.60} & 3.20          & \textbf{3.60} & \textbf{2.60} & \textbf{5.60} & \textbf{24.20} & \textbf{10.00} & \textbf{23.10} & \textbf{3.20} & 1.90          & \textbf{2.00} & \textbf{16.10} & \textbf{7.70} & \textbf{8.00} \\
            \bottomrule
        \end{tabular}
    }
    \caption{\textbf{Quantitative results on SemanticKITTI \texttt{test}.} $^\dagger$ denotes the results provided by \cite{MonoScene}. $^\ast$ represents the reproduced results in ~\cite{TPVFormer, OccFormer}. The best results are in \textbf{bold}.}
    \label{tab:sem_kitti_test}
\end{table*}

%% file: tab/2_kitti_360.tex
\begin{table*}[ht]
    \centering
    \newcommand{\clsname}[2]{
        \rotatebox{90}{
            \hspace{-6pt}
            \textcolor{#2}{$\blacksquare$}
            \hspace{-6pt}
            \renewcommand\arraystretch{0.6}
            \begin{tabular}{l}
                #1                                       \\
                \hspace{-4pt} ~\tiny(\sscbkitfreq{#2}\%) \\
            \end{tabular}
        }}
    \newcommand{\empa}[1]{\textbf{#1}}
    \newcommand{\empb}[1]{\underline{#1}}
    \renewcommand{\tabcolsep}{2pt}
    \renewcommand\arraystretch{1.2}
    \scalebox{0.78}
    {
        \begin{tabular}{l|rrr>{\columncolor{gray!20}}r|rrrrrrrrrrrrrrrrrr}
            \toprule
            \multicolumn{1}{c|}{Method}                                 &
            \multicolumn{1}{c}{IoU}                                     &
            \multicolumn{1}{c}{Prec.}                                   &
            \multicolumn{1}{c}{Rec.}                                    &
            mIoU                                                        &
            \multicolumn{1}{c}{\clsname{car}{car}}                      &
            \multicolumn{1}{c}{\clsname{bicycle}{bicycle}}              &
            \multicolumn{1}{c}{\clsname{motorcycle}{motorcycle}}        &
            \multicolumn{1}{c}{\clsname{truck}{truck}}                  &
            \multicolumn{1}{c}{\clsname{other-veh.}{othervehicle}}      &
            \multicolumn{1}{c}{\clsname{person}{person}}                &
            \multicolumn{1}{c}{\clsname{road}{road}}                    &
            \multicolumn{1}{c}{\clsname{parking}{parking}}              &
            \multicolumn{1}{c}{\clsname{sidewalk}{sidewalk}}            &
            \multicolumn{1}{c}{\clsname{other-grnd.}{otherground}}      &
            \multicolumn{1}{c}{\clsname{building}{building}}            &
            \multicolumn{1}{c}{\clsname{fence}{fence}}                  &
            \multicolumn{1}{c}{\clsname{vegetation}{vegetation}}        &
            \multicolumn{1}{c}{\clsname{terrain}{terrain}}              &
            \multicolumn{1}{c}{\clsname{pole}{pole}}                    &
            \multicolumn{1}{c}{\clsname{traf.-sign}{trafficsign}}       &
            \multicolumn{1}{c}{\clsname{other-struct.}{otherstructure}} &
            \multicolumn{1}{c}{\clsname{other-obj.}{otherobject}}
            \\
            \midrule
            \multicolumn{23}{l}{\textit{LiDAR-based methods}}                                                                                                                                                                                                                                                                                                                                               \\
            \hline
            SSCNet~\cite{SSCNet}                                        & \empa{53.58} & 69.63        & \empa{69.92} & 16.95        & \empa{31.95} & 0.00        & 0.17        & 10.29        & 0.00         & 0.07        & \empa{65.70} & \empa{17.33} & \empa{41.24} & 3.22        & \empa{44.41} & 6.77        & \empa{43.72} & \empa{28.87} & 0.78         & 0.75        & 8.69         & 0.67         \\
            LMSCNet~\cite{LMSCNet}                                      & 47.35        & \empa{72.77} & 57.55        & 13.65        & 20.91        & 0.00        & 0.00        & 0.26         & 0.58         & 0.00        & 62.95        & 13.51        & 33.51        & 0.20        & 43.67        & 0.33        & 40.01        & 26.80        & 0.00         & 0.00        & 3.63         & 0.00         \\
            \specialrule{0.7pt}{0pt}{0pt}
            \multicolumn{23}{l}{\textit{Camera-based methods}}                                                                                                                                                                                                                                                                                                                                              \\
            \hline
            MonoScene~\cite{MonoScene}                                  & 37.87        & 56.73        & 53.26        & 12.31        & 19.34        & 0.43        & 0.58        & 8.02         & 2.03         & 0.86        & 48.35        & 11.38        & 28.13        & 3.32        & 32.89        & 3.53        & 26.15        & 16.75        & 6.92         & 5.67        & 4.20         & 3.09         \\
            TPVFormer~\cite{TPVFormer}                                  & 40.22        & 59.32        & \empb{55.54} & 13.64        & 21.56        & 1.09        & 1.37        & 8.06         & 2.57         & 2.38        & 52.99        & 11.99        & 31.07        & 3.78        & 34.83        & 4.80        & 30.08        & 17.52        & 7.46         & 5.86        & 5.48         & 2.70         \\
            VoxFormer~\cite{VoxFormer}                                  & 38.76        & 58.52        & 53.44        & 11.91        & 17.84        & 1.16        & 0.89        & 4.56         & 2.06         & 1.63        & 47.01        & 9.67         & 27.21        & 2.89        & 31.18        & 4.97        & 28.99        & 14.69        & 6.51         & 6.92        & 3.79         & 2.43         \\
            OccFormer~\cite{OccFormer}                                  & 40.27        & 59.70        & 55.31        & 13.81        & 22.58        & 0.66        & 0.26        & 9.89         & 3.82         & 2.77        & 54.30        & 13.44        & 31.53        & 3.55        & \empb{36.42} & 4.80        & 31.00        & \empb{19.51} & 7.77         & 8.51        & 6.95         & 4.60         \\
            \hline
            \name                                                       & \empb{44.12} & \empb{69.24} & 54.88        & \empa{18.58} & \empb{30.02} & \empa{1.85} & \empa{5.90} & \empa{25.07} & \empa{12.06} & \empa{8.20} & \empb{54.94} & \empb{13.83} & \empb{32.76} & \empa{6.93} & 35.11        & \empa{8.58} & \empb{38.33} & 11.52        & \empa{14.01} & \empa{9.57} & \empa{14.44} & \empa{11.28} \\
            \bottomrule
        \end{tabular}
    }
    \caption{\textbf{Quantitative results on SSCBench-KITTI360 \texttt{test}.} The results for counterparts are provided in \cite{SSCBench}. The best results among all methods are in \empa{bold}, and the best results for camera-based methods are \empb{underlined}.}
    \label{tab:kitti_360_test}
\end{table*}

%% file: tab/3_ablat_mods.tex


\begin{table*}[ht]
    \centering
    \newcommand{\gray}[1]{\textcolor{gray}{#1}}
    {
        \begin{tabular}{l|l>{\columncolor{gray!20}}lcc}
            \toprule

            Method                 & \multicolumn{1}{c}{IoU}       & \multicolumn{1}{>{\columncolor{gray!20}}c}{mIoU} & Params (M) & FLOPs (G) \\
            \midrule
            Baseline               & 34.06                         & 10.44                                            & 57.22      & 529.20    \\
            + Pre-trained Encoder  & 35.97 \gray{(+1.91)}          & 12.59 \gray{(+2.15)}                             & 57.22      & 529.20    \\
            + Voxel Proposal Layer & 36.54 \gray{(+0.57)}          & 13.34 \gray{(+0.75)}                             & 57.42      & 535.84    \\
            \hline
            + Symphonies Decoder   & \textbf{41.92} \gray{(+5.38)} & \textbf{14.89} \gray{(+1.55)}                    & 59.31      & 611.89    \\
            \bottomrule
        \end{tabular}
    }
    \caption{\textbf{Ablation study on architectural components in Symphonies.}}
    \label{tab:ablat_mods}
\end{table*}

%% file: tab/4_ablat_decoder.tex
\begin{table}[ht]
    \centering
    \renewcommand{\tabcolsep}{4pt}
    \renewcommand\arraystretch{1.1}
    \scalebox{0.92}{
        \begin{tabular}{ccccc|c>{\columncolor{gray!20}}c}
            \toprule
            Scn.       & Scn.-Inst. & Inst.-Img. & Inst.-Scn. & Inst.      &                       & \cellcolor{gray!20}                       \\
            SA         & CA         & CA         & CA         & SA         & \multirow{-2}{*}{IoU} & \multirow{-2}{*}{\cellcolor{gray!20}mIoU} \\
            \midrule
                       &            &            &            &            & 35.97                 & 13.34                                     \\
            \checkmark &            &            &            &            & 41.18                 & 14.01                                     \\
            \checkmark & \checkmark & \checkmark &            &            & 41.14                 & 14.56                                     \\
            \checkmark & \checkmark &            & \checkmark &            & 41.26                 & 14.66                                     \\
            \checkmark & \checkmark &            & \checkmark & \checkmark & 41.78                 & 14.76                                     \\
            \hline
            \checkmark & \checkmark & \checkmark & \checkmark & \checkmark & \textbf{41.92}        & \textbf{14.89}                            \\
            \bottomrule
        \end{tabular}}
    \caption{\textbf{Ablation study on Symphonies Decoder.}}
    \label{tab:ablat_attns}
\end{table}

%% file: tab/5_ablat_vpl.tex
\begin{table}[ht]
        \centering
        \begin{tabular}{l|c>{\columncolor{gray!20}}c}
            \toprule
            2D-to-3D Projection \  & IoU            & mIoU           \\
            \midrule
            FLoSP~\cite{MonoScene} & 36.02          & 11.96          \\
            VPL (mono)             & 38.37          & 12.20          \\
            VPL (stereo)           & \textbf{41.92} & \textbf{14.89} \\
            \bottomrule
        \end{tabular}
        \caption{\textbf{Ablation on the Depth-Rectified Voxel Proposal Layer.}}
        \label{tab:ablat_vpl}
\end{table}

%% file: tab/6_ablat_encoder.tex
\begin{table}[ht]
        \centering
        \begin{tabular}{cc|c>{\columncolor{gray!20}}c}
            \toprule
            Pre-trained Encoder & Inst. Queries & IoU            & mIoU           \\
            \midrule
                                &               & 41.09          & 13.53          \\
                                & \checkmark    & 41.42          & 13.32          \\
            \checkmark          &               & 41.18          & 14.01          \\
            \hline
            \checkmark          & \checkmark    & \textbf{41.92} & \textbf{14.89} \\
            \bottomrule
        \end{tabular}
        \caption{\textbf{Ablation study on the Instance-Aware Image Encoder.}}
        \label{tab:ablat_encoder}
\end{table}

%% file: tab/8_sem_kitti_val.tex
\begin{table*}[htb]
    \centering
    \newcommand{\clsname}[2]{
        \rotatebox{90}{
            \hspace{-6pt}
            \textcolor{#2}{$\blacksquare$}
            \hspace{-6pt}
            \renewcommand\arraystretch{0.6}
            \begin{tabular}{l}
                #1                                      \\
                \hspace{-4pt} ~\tiny(\semkitfreq{#2}\%) \\
            \end{tabular}
        }}
    \renewcommand{\tabcolsep}{2pt}
    \renewcommand\arraystretch{1.1}
    \scalebox{0.84}
    {
        \begin{tabular}{l|r>{\columncolor{gray!20}}r|rrrrrrrrrrrrrrrrrrrr}
            \toprule
            Method                               &
            \multicolumn{1}{c}{IoU}              &
            mIoU                                 &
            \clsname{road}{road}                 &
            \clsname{sidewalk}{sidewalk}         &
            \clsname{parking}{parking}           &
            \clsname{other-grnd.}{otherground}   &
            \clsname{building}{building}         &
            \clsname{car}{car}                   &
            \clsname{truck}{truck}               &
            \clsname{bicycle}{bicycle}           &
            \clsname{motorcycle}{motorcycle}     &
            \clsname{other-veh.}{othervehicle}   &
            \clsname{vegetation}{vegetation}     &
            \clsname{trunk}{trunk}               &
            \clsname{terrain}{terrain}           &
            \clsname{person}{person}             &
            \clsname{bicyclist}{bicyclist}       &
            \clsname{motorcyclist}{motorcyclist} &
            \clsname{fence}{fence}               &
            \clsname{pole}{pole}                 &
            \clsname{traf.-sign}{trafficsign}
            \\
            \midrule
            LMSCNet$^\dagger$~\cite{LMSCNet}     & 28.61          & 6.70           & 40.68          & 18.22          & 4.38           & 0.00           & 10.31          & 18.33          & 0.00           & 0.00          & 0.00          & 0.00           & 13.66          & 0.02           & 20.54          & 0.00          & 0.00          & 0.00          & 1.21          & 0.00          & 0.00          \\
            AICNet$^\dagger$~\cite{AICNet}       & 29.59          & 8.31           & 43.55          & 20.55          & 11.97          & 0.07           & 12.94          & 14.71          & 4.53           & 0.00          & 0.00          & 0.00           & 15.37          & 2.90           & 28.71          & 0.00          & 0.00          & 0.00          & 2.52          & 0.06          & 0.00          \\
            JS3C-Net$^\dagger$~\cite{JS3CNet}    & 38.98          & 10.31          & 50.49          & 23.74          & 11.94          & 0.07           & 15.03          & 24.65          & 4.41           & 0.00          & 0.00          & 6.15           & 18.11          & 4.33           & 26.86          & 0.67          & 0.27          & \textbf{0.20} & 3.94          & 3.77          & 1.45          \\
            MonoScene$^\ast$~\cite{MonoScene}    & 36.86          & 11.08          & 56.52          & 26.72          & 14.27          & 0.46           & 14.09          & 23.26          & 6.98           & 0.61          & 0.45          & 1.48           & 17.89          & 2.81           & 29.64          & 1.86          & 1.20          & 0.00          & 5.84          & 4.14          & 2.25          \\
            TPVFormer~\cite{TPVFormer}           & 35.61          & 11.36          & 56.50          & 25.87          & 20.60          & 0.85           & 13.88          & 23.81          & 8.08           & 0.36          & 0.05          & 4.35           & 16.92          & 2.26           & 30.38          & 0.51          & 0.89          & 0.00          & 5.94          & 3.14          & 1.52          \\
            VoxFormer~\cite{VoxFormer}           & \textbf{44.02} & 12.35          & 54.76          & 26.35          & 15.50          & 0.70           & 17.65          & 25.79          & 5.63           & 0.59          & 0.51          & 3.77           & 24.39          & 5.08           & 29.96          & 1.78          & \textbf{3.32} & 0.00          & 7.64          & 7.11          & 4.18          \\
            OccFormer~\cite{OccFormer}           & 36.50          & 13.46          & 58.85          & 26.88          & 19.61          & 0.31           & 14.40          & 25.09          & \textbf{25.53} & 0.81          & 1.19          & 8.52           & 19.63          & 3.93           & \textbf{32.62} & 2.78          & 2.82          & 0.00          & 5.61          & 4.26          & 2.86          \\
            NDC-Scene~\cite{NDCScene}            & 37.24          & 12.70          & \textbf{59.20} & \textbf{28.24} & \textbf{21.42} & \textbf{1.67}  & 14.94          & 26.26          & 14.75          & 1.67          & 2.37          & 7.73           & 19.09          & 3.51           & 31.04          & 3.60          & 2.74          & 0.00          & 6.65          & 4.53          & 2.73          \\
            \hline
            \name{}                              & 41.92          & \textbf{14.89} & 56.37          & 27.58          & 15.28          & 0.95           & \textbf{21.64} & \textbf{28.68} & 20.44          & \textbf{2.54} & \textbf{2.82} & \textbf{13.89} & \textbf{25.72} & \textbf{6.60} & 30.87           & \textbf{3.52} & 2.24          & 0.00          & \textbf{8.40} & \textbf{9.57} & \textbf{5.76} \\
            \bottomrule
        \end{tabular}
    }
    \caption{\textbf{Quantitative results on SemanticKITTI \texttt{val}.} $^\dagger$ denotes the results provided by \cite{MonoScene}. $^\ast$ represents the reproduced results in ~\cite{TPVFormer, OccFormer}. The best results are in \textbf{bold}.}
    \label{tab:sem_kitti_val}
\end{table*}

%% file: sec/5_conclusion.tex
\section{Conclusion}
\label{sec:conclusion}

In this paper, we introduced Symphonies, a novel paradigm for 3D Semantic Scene Completion. Symphonies effectively integrates instance-centric semantics and scene context from both images and volumes, addressing the limitations posed by geometric ambiguity in prior voxel-wise modeling methods.
Extensive experiments demonstrate the superiority of our approach over existing methods. We anticipate Symphonies to inspire future research and contribute to advancements in autonomous driving and 3D perception.

\begin{figure}
    \centering
    \includegraphics[width=\linewidth ]{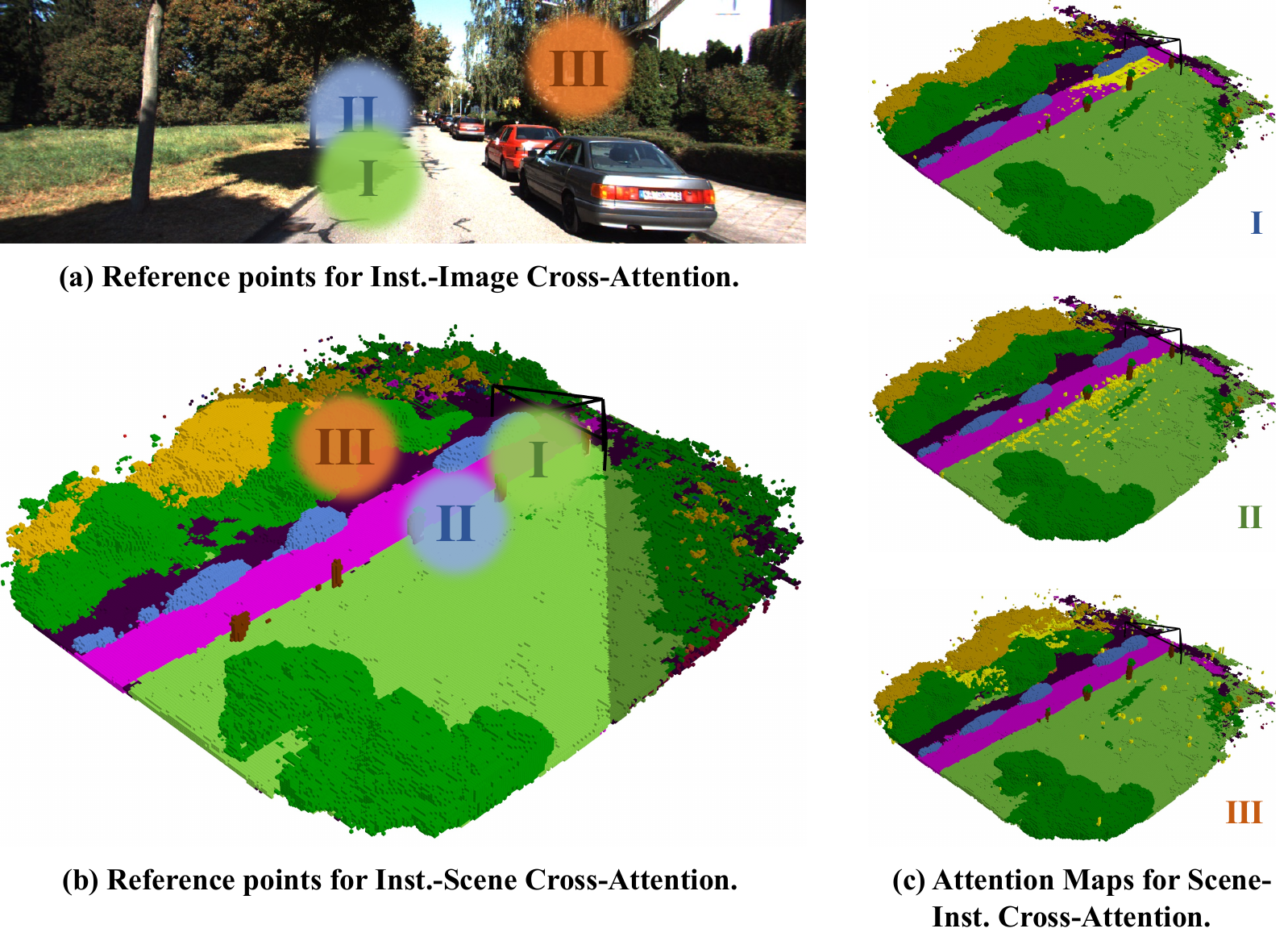}
    \caption{\textbf{Analysis of attention maps within \name{}.}}
    \label{fig:vis_attn}
\end{figure}

\paragraph{Broader Impacts and Limitations.}
We envision that the incorporation of instance-centric representations in our paradigm will contribute to future research, especially within the context of end-to-end paradigm in UniAD~\cite{UniAD} and spur autonomous driving. However, it is essential to acknowledge that our work represents an initial step in leveraging instance-centric queries. Limitations, such as the absence of instance-level annotations, may affect the performance of instance-based methods. Our future endeavors aim to extend this paradigm to multi-view and temporal scenarios. Despite these limitations, we believe that our paradigm holds promise in advancing the field of SSC.